\documentclass{article}

\usepackage[final,nonatbib]{nips_2018}

\usepackage[utf8]{inputenc} 
\usepackage[T1]{fontenc}    
\usepackage{hyperref}       
\usepackage{url}            
\usepackage{booktabs}       
\usepackage{amsfonts}       
\usepackage{amsmath}
\usepackage{nicefrac}       
\usepackage{microtype}      
\usepackage{epsfig}
\usepackage{epstopdf}
\usepackage{subcaption}
\usepackage{dsfont}
\usepackage[english]{babel}
\usepackage[numbers]{natbib}
\usepackage{graphicx}

\DeclareMathOperator*{\argmax}{arg\,max}

\title{Trajectory-based Learning for Ball-in-Maze Games}

\author{
  Sujoy Paul\thanks{Work done as an intern at Mitsubishi Electric Research Laboratories (MERL), USA} \\
  University of California--Riverside \\
  Riverside, CA 92521 \\
  \texttt{supaul@ece.ucr.edu} \\
  \And
  Jeroen van Baar \\
  Mitsubishi Electric Research Laboratories (MERL) \\
  Cambridge, MA 02139 \\
  \texttt{jeroen@merl.com} \\
}

\begin{document}

\maketitle

\begin{abstract}
Deep Reinforcement Learning has shown tremendous success in solving several games and tasks in robotics. However, unlike humans, it generally requires a lot of training instances. Trajectories imitating to solve the task at hand can help to increase sample-efficiency of deep RL methods. In this paper, we present a simple approach to use such trajectories, applied to the challenging Ball-in-Maze Games, recently introduced in the literature. We show that in spite of not using human-generated trajectories and just using the simulator as a model to generate a limited number of trajectories, we can get a speed-up of about 2-3x in the learning process. We also discuss some challenges we observed while using trajectory-based learning for very sparse reward functions.
\end{abstract}

\section{Introduction}

\begin{figure*}[!t]
	\begin{center}
		\begin{subfigure}{0.48\textwidth}
			\vspace{3mm}
			\includegraphics[scale=0.39]{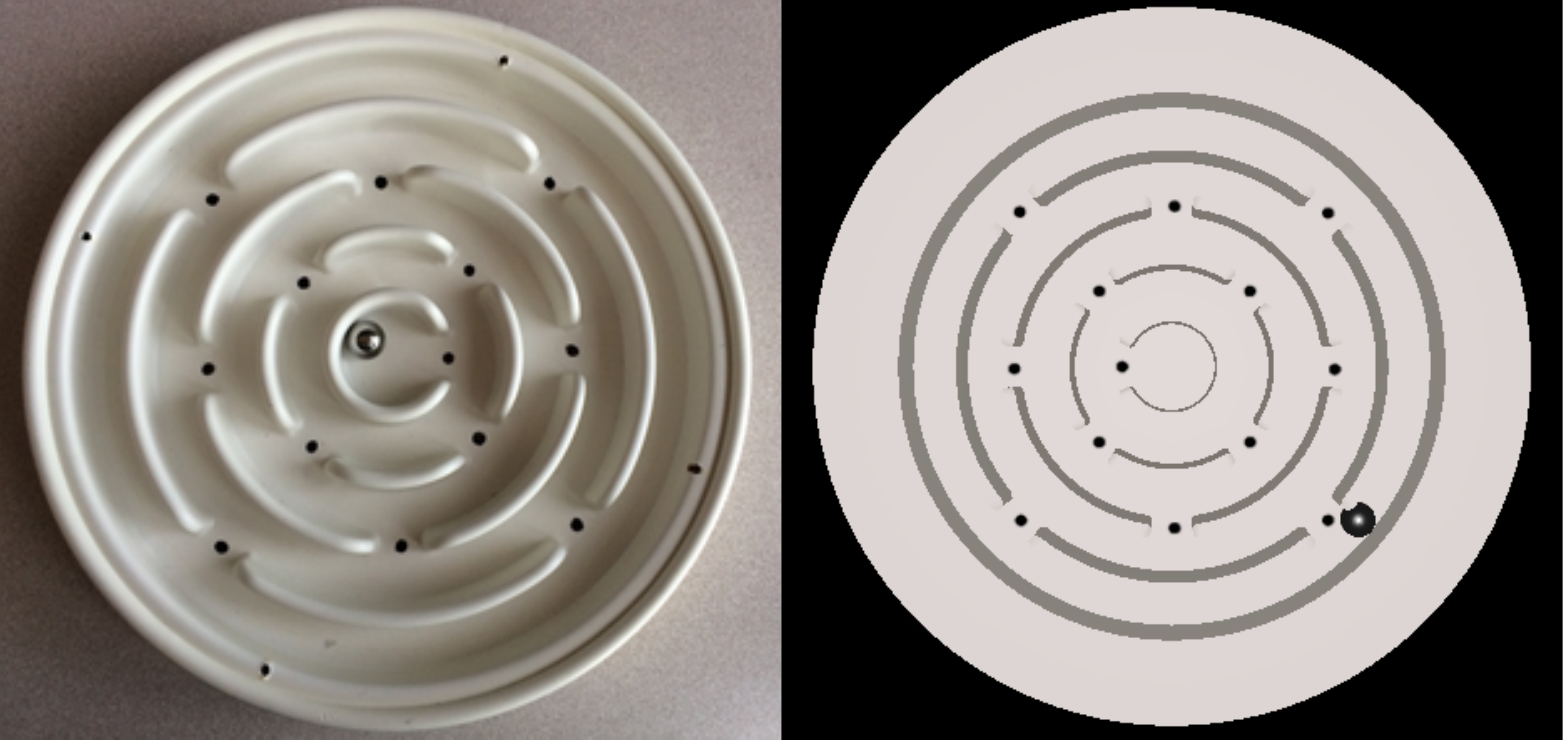}
			\vspace{1.5mm}
			\caption{}			
			\label{fig:mazes}
		\end{subfigure}
		\begin{subfigure}{0.48\textwidth}
			\includegraphics[scale=0.15]{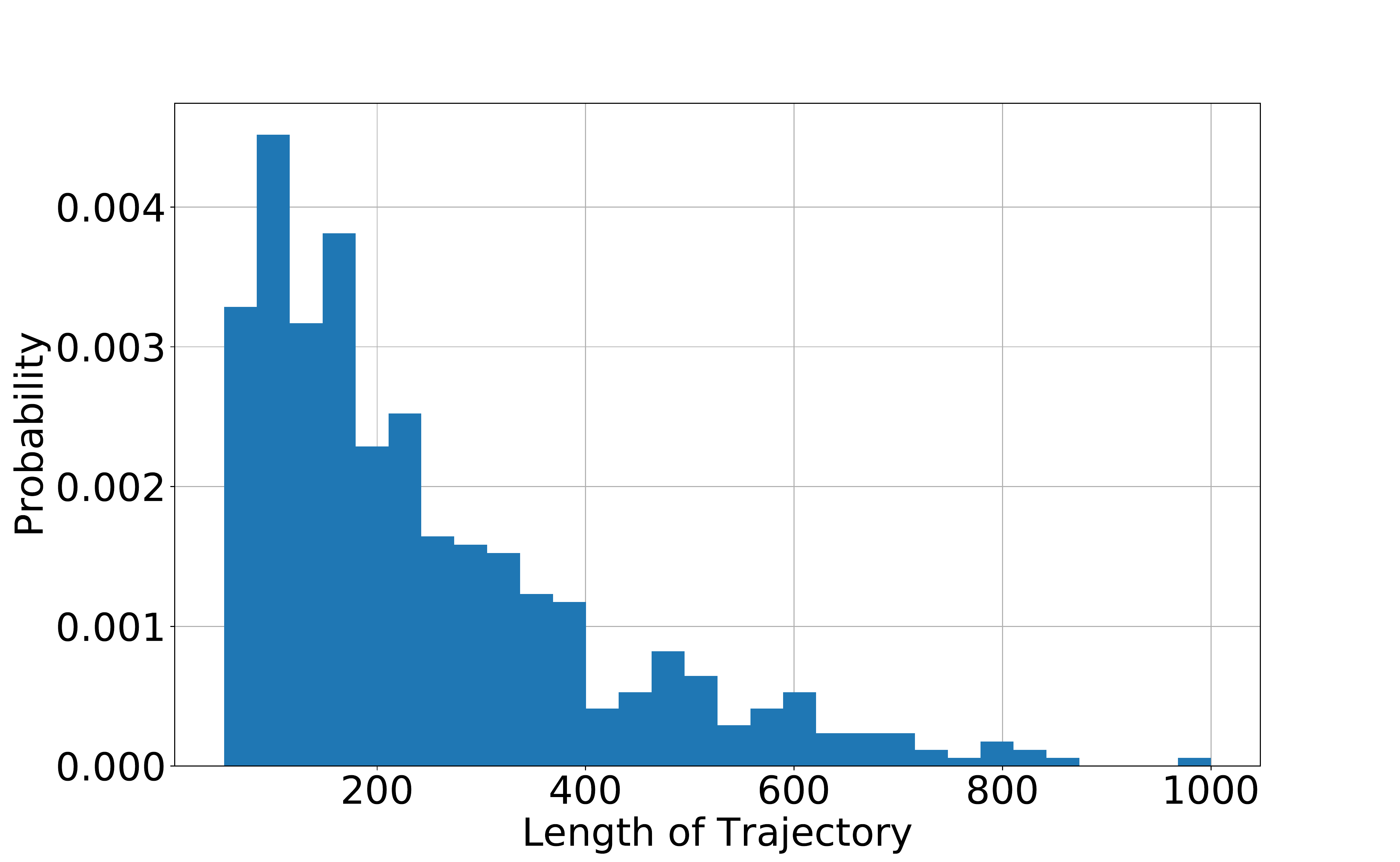}
			\vspace{-2mm}
			\caption{}
			\label{fig:distribution}
		\end{subfigure}
		\caption{(a) Ball-in-Maze Game puzzle: the real (\textbf{Left}) and simulated (\textbf{Right}) BiMGame. (b) This plot presents the distribution of trajectory lengths generated by using the simulator as the model }
		\label{fig:im}
	\end{center}
\vspace*{-3mm}
\end{figure*}

The challenging task of solving a Ball-in-Maze Game (BiMGame) using robotics has recently been introduced \cite{vanbaar:2018:simtoreal, jharomeres:2018:modelbasedmaze} (see Fig. \ref{fig:mazes}). This task is challenging since maze puzzles exhibit complex dynamics due to static friction, collisions with the geometry (or between marbles in the case of more than one marble), and long-horizon planning. Sample-efficient, model-based Reinforcement Learning (RL) approaches, e.g.,~\cite{levine2016end}, are desirable. However, for BiMGames, this is not solved~\cite{jharomeres:2018:modelbasedmaze}. Instead, in \cite{vanbaar:2018:simtoreal} the authors propose to use model-free deep RL \cite{mnih:2016:A3C}, and demonstrated a transfer learning approach from simulation to a real robot.

A major drawback of model-free RL is the lack of sample efficiency since exploration is performed in an unstructured fashion. In this paper, we investigate several algorithms to reduce the required number of samples. We adopt ideas from trajectory-based imitation learning methods, in which controllers are learned to model the distribution of trajectories and are subsequently refined. The necessary trajectories are provided by an expert, often a human. Rather than requiring a human expert to generate trajectories for each instance of this game, in the context of Sim2Real, we can exploit the simulator to generate those trajectories instead. We use these trajectories to pre-train the network policy and then fine-tune it with policy gradients based RL.

%
In the remainder of this paper, we first briefly discuss previously introduced Imitation Learning (IL) approaches along with the challenges we encountered to learn BiMGame. Thereafter, we present the details of the trajectory generation procedure, followed by the IL-RL framework we take to learn from these trajectories and the experimental results for different tasks of BiMGame. We conclude the paper with the challenges we faced with the different tasks and future directions to address them.

%
%
%

\section{Related Work}
\label{sec:related}

Recently, Deep Neural Networks (DNN) has enabled RL approaches to achieve impressive results. It's beyond the scope of this paper to list all prior work, and we only discuss the most relevant. The Sim2Real approach in~\cite{vanbaar:2018:simtoreal} is based on Asynchronous Advantage Actor-Critic (A3C)~\cite{mnih:2016:A3C}. However, deep RL learns on a trial-and-error basis, thus requiring a significant number of training samples.

Many imitation learning approaches use trajectories obtained by experts~\cite{ross2011reduction, ross2014reinforcement, sun2017deeply,cheng2018fast} to guide the learning process to be more structured. Of particular interest is the AggreVateD method~\cite{sun2017deeply}, which is similar to A3C, with the exception that the advantage estimate is that of an expert policy rather than the current policy being learned. We found that this approach relies heavily on very accurate estimation of the value function at all states visited and turned out to not solve BiMGame consistently--even with the expert value function being that of a trained policy, which is able to solve the game consistently.

To make RL sample efficient, the authors in ~\cite{nagabandi2017neural} propose to start with model-based deep RL, followed by model-free fine-tuning. The cloning from model-based to model-free is performed by DAgger \cite{ross2011reduction}, using the trajectories generated by model-based RL. As the model is learned from data, the performance obtained by the model-based method is quite low. Modeling even more complex dynamics and high dimensional state-space such as in BiMGame would be difficult to achieve. The authors in~\cite{rahmatizadeh2016learning} propose a Sim2Real approach which avoids RL, and instead rely on supervised training from expert trajectories. Compared to BiMGame, the dynamics of these tasks are much simpler.


\section{Generating Trajectories: Using the Simulator as the Model}
\label{sec:method}

The trajectories to be used for learning can be obtained from a human teacher \cite{rahmatizadeh2016learning,yu2018one} or from planning based algorithms \cite{guo2014deep}. In \cite{nagabandi2017neural}, the authors first learned a neural network dynamics model and then used it to generate trajectories. However, when learning in the simulator (in a Sim2Real paradigm), we could leverage upon the internal physics engine to roll-out in time and generate trajectories by optimizing the cumulative reward function. Formally, at time step $t$, we aim to solve the following:
\begin{equation}
a^*_{t:{t+H}} = \argmax_{a_{t:{t+H}}} \sum_{t'=t}^{t+H-1} r(s_{t'},a_{t'},s_{t'+1}), \ \ \ \ \text{s.t.}, s_{t'+1}=\mathcal{S}(s_t',a_t')
\label{gen}
\end{equation}
where $\mathcal{S}$ is the simulator, $r$ is the reward function, $H$ is the horizon of optimization and $a_{t:{t+H}}$ is the set of actions from $a_t$ to $a_{t+H}$. We only take the first action $a^*_t$, move to state $s_{t+1}$ and repeat the same optimization procedure to choose the next action. As we use a non-differentiable simulator, we employ a random shooting strategy \cite{rao2009survey} where we sample $K$ sets of $a^*_{t:{t+H}}$ and choose the one which maximizes the object function. We set $K=10$ empirically.

Using this method, we obtain a dataset of trajectories $\mathcal{D}=\{\{(s_{ti},a^*_{ti}, r_{ti})\}_{t=1}^{n_i}\}_{i=1}^{n_t}$, each of which is a sequence of state-action-reward triplets. It may be noted that the optimization process and thus the trajectories obtained using the above process is not optimal. Moreover, the choice of reward function used in the above optimization procedure plays an important role towards the error in the trajectories. In this work, we use the following distance based reward function:
\begin{equation}
r(s_t,a_t,s_{t+1}) = d(s_{t+1}) - d(s_t)
\label{reward}
\end{equation}
where $d(s_t)$ is the radial distance of the ball at time $t$ from the center of the board. Note that as the radial path is not a feasible one due to obstructions,  this reward function is only good near the gate openings and not otherwise. In spite of using such a reward function, the above method is able to solve the BiMGame consistently. Fig. \ref{fig:distribution} shows a distribution of trajectory lengths required to solve the BiMGame using the above method. We next describe the methods we use to learn the tasks using the trajectories $\mathcal{D}$. The methods are agnostic to the procedure with which we obtain the trajectories.

\section{Learning From Trajectories}
\subsection{Supervised Pre-training}
The dataset $\mathcal{D}$ forms a rich source of information to guide the RL agents for faster convergence. We use this set of trajectories to pre-train the DNN policy. We follow the architecture structure of A3C \cite{mnih:2016:A3C} with consists of a DNN with two heads - one for policy $\pi_\theta(a_t|s_t)$ and the other one for value estimation $V_\phi(s_t)$, with partially shared parameters between $\theta$ and $\phi$. We use two loss functions to pre-train the deep neural network as follows:
\begin{equation}
\mathcal{L} = \sum_{i=1}^{n_t} \sum_{t=1}^{n_i} \Big[\sum_{c=1}^C a^{*c}_t\log \pi_\theta(a_{ti}^c|s_{ti}) + \frac{1}{2}(V_\phi(s_{ti})-\sum_{t'=t}^{n_i}\gamma r_{ti})^2 \Big] + ||\theta \cup \phi||_2^2
\label{loss}
\end{equation}
where $C$ is the number of discrete actions. The first part of the loss function is the cross-entropy loss to train the policy network, the second part of the loss is to train the value function estimator and the last part is the  $l2$ regularization loss. We denote the policy learned by minimizing Eqn. \ref{loss} as $\pi^s$.

$\pi^s$ possess the ability to take actions with low error rates at the states sampled from the distribution induced by $\mathcal{D}$. However, a small error at the beginning would compound quadratically \cite{ross2011reduction} with time as the DNN agent starts visiting states which are not sampled from the distribution of $\mathcal{D}$. Algorithms like DAgger can be used to finetune the policy on the states distributed by rolling out $\pi^s$, by using ground-truth labels obtained from an expert policy $\pi^*$. This query to $\pi^*$ is often very costly and even may not be feasible in some applications.

In this paper, instead of taking the route of DAgger, we fine-tune the policy $\pi^s$ using policy gradient RL in an A3C framework. This method is quite simple as we just need to initialize A3C with $\pi^s$ instead of randomly initialized network parameters.

\subsection{Value Function as a Reward}
Another way of using the trajectories for sample-efficient learning is to shape the reward \cite{ng1999policy} using the value function learned from the trajectories. We train a network with only the value head to estimate $\hat{V}_\phi(s_t)$ using the dataset $\mathcal{D}$ by optimizing only the last two parts of Eqn. \ref{loss}. Thereafter we use A3C but with a transformed reward function as follows:
\begin{equation}
\bar{r}(s_t,a_t,s_{t+1})=r(s_t,a_t,s_{t+1}) + \gamma \hat{V}_\phi(s_{t+1}) - \hat{V}_\phi(s_t)
\end{equation}
While using this reward function, the A3C agents can either start from randomly initialized weights or from weights pre-trained by minimizing the loss in Eqn. \ref{loss}. Please note that for the latter, $V_\phi$ and $\hat{V}_\phi$ represent two different networks. $\hat{V}_\phi$ is kept fixed after training and just serves as a source of an auxiliary reward function and $V_\phi$ is finetuned using the A3C framework.

\section{Experiments}
\label{sec:experiments}

\textbf{Tasks.} We perform experiments on three instances of BiMGame. All the games are initialized with the marble being at the outermost ring. We refer to the first instance as FULL, where the agent receives a +/- 1 reward for moving a marble through a gate towards/away from the center. The goal is to move the marble into the central portion of the maze. We define additional instances with a more sparse reward signal, where a reward of +1 is only received if the goal (terminal) is attained, otherwise the reward is zero. We refer to these games as Steps-to-Go (STG). We define STG1 as the goal for the marble to move through one of the gates between the first and second ring, and STG2 as the goal of moving through one of the gates between the second and third ring. There are 5 possible actions in the games - clockwise and anti-clockwise rotations of $1^{\circ}$ along the two principal axes on the plane of the board and a \textit{No-Operation} action.

\textbf{Algorithms.} We compare the following algorithms: A3C, Supervised Pre-training followed by A3C fine-tuning, A3C with Value based reward function and Supervised Pre-training followed by A3C fine-tuning with Value based reward function. We also compare with DAgger \cite{ross2011reduction}.

\textbf{Deep Neural Network.}The network architecture we train is Conv-Conv-FC-LSTM. The input to the LSTM is the previous layer feature along with the previous step action and reward. After the LSTM layer, we add two FC heads for the policy and value function.

\begin{figure*}[!t]
	\begin{center}
		\hspace{-10mm}
		\begin{subfigure}{0.33\textwidth}
			\includegraphics[scale=0.126]{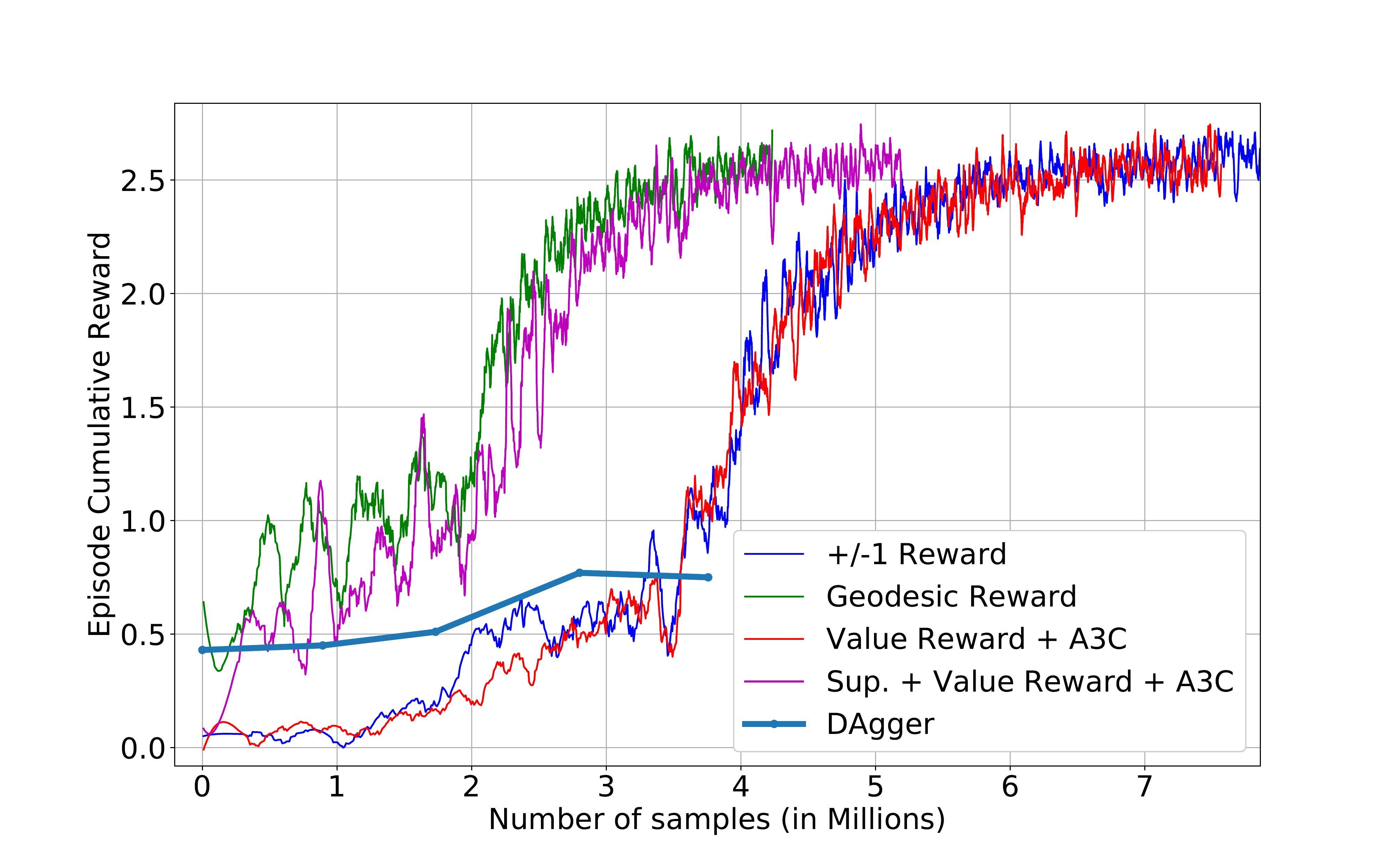}
			\caption{FULL}			
			\label{fig:original}
		\end{subfigure}
		\begin{subfigure}{0.33\textwidth}
			\includegraphics[scale=0.126]{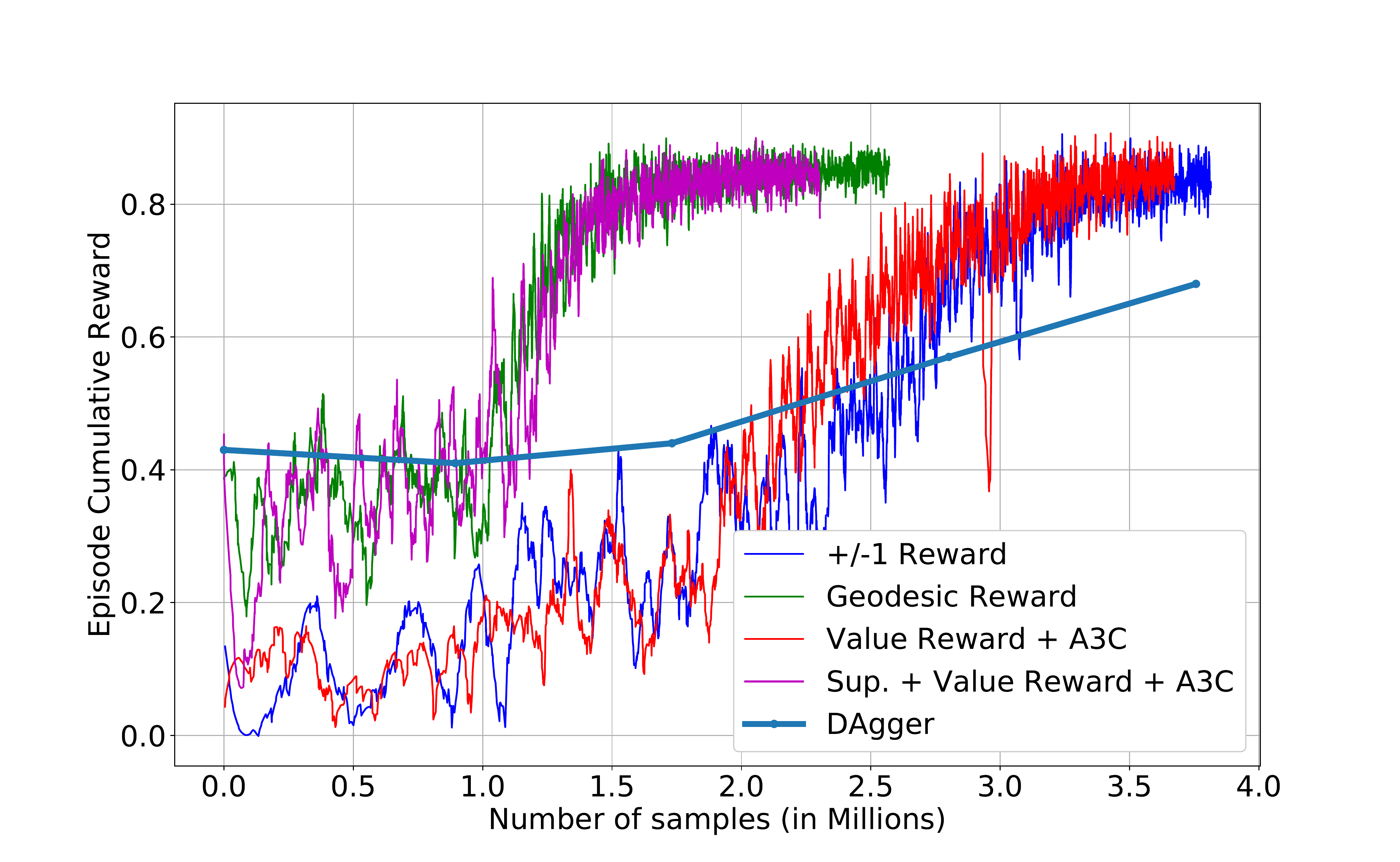}
			\caption{STG1}
			\label{fig:stg1}
		\end{subfigure}
		\begin{subfigure}{0.33\textwidth}
			\includegraphics[scale=0.127]{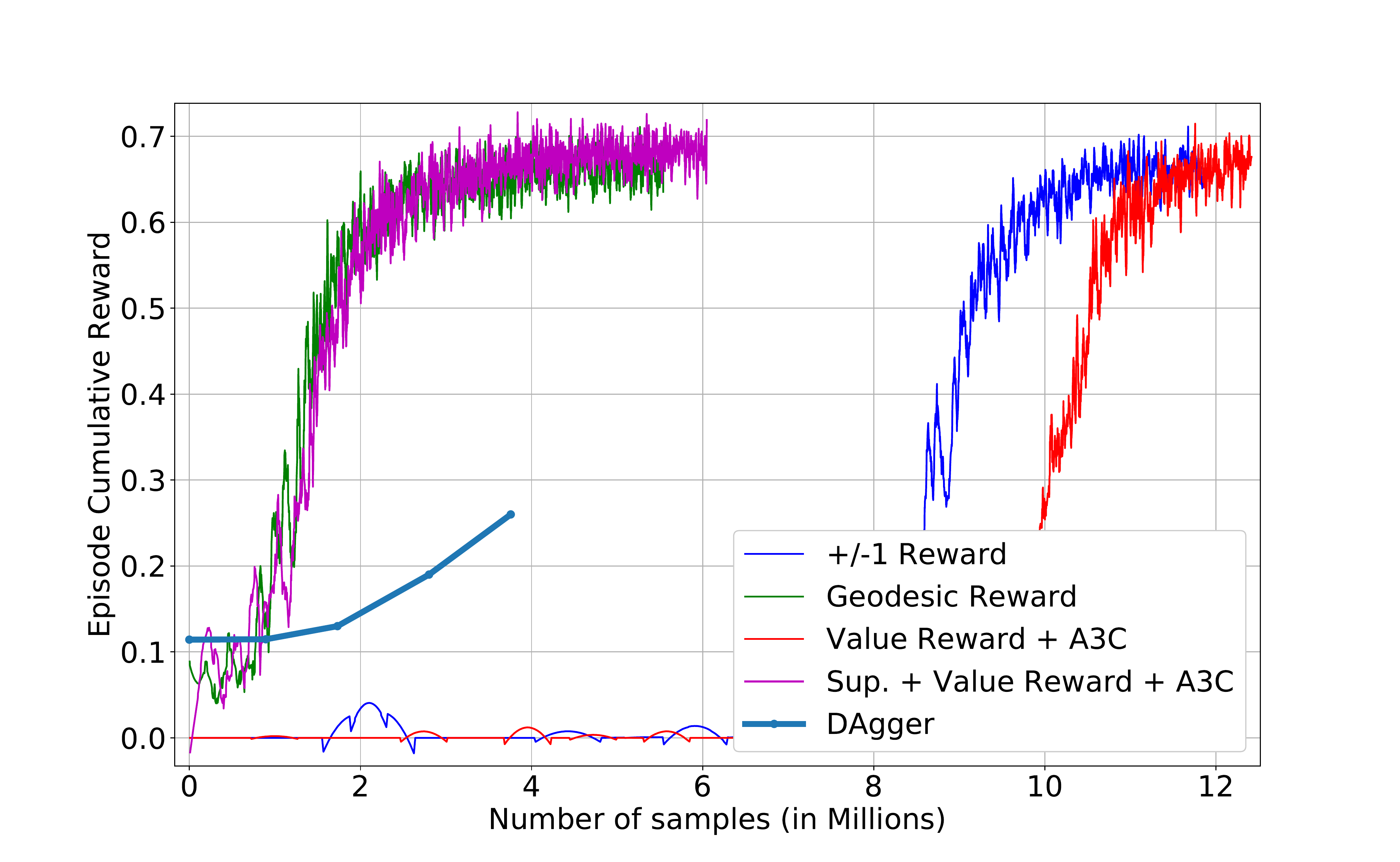}
			\caption{STG2}
			\label{fig:stg2}
		\end{subfigure}
	   \vspace*{-1mm}
		\caption{This presents the learning curves for the three tasks (a) FULL, (b) STG1 and (c) STG2. }
		\label{fig:plots}
	\end{center}
	\vspace{-4mm}
\end{figure*}

\textbf{Results.}
During pre-training, we obtain a maximum classification accuracy of $27.5\%$ over a test set of trajectories. The low accuracy shows that the problem at hand is quite difficult to learn just using the trajectories and supervised training. This result is in contrast to that in \cite{nagabandi2017neural}, where the authors achieved a perfect cloning from model-based to model-free for their tasks using trajectories together with DAgger, which requires a step to query an expert and can be costly in general. Also, it can be observed in the plots that DAgger perform much worse in the harder tasks such as STG2 and FULL compared to the easier task of STG1. Moreover, DAgger based imitation learning performs much worse than the framework presented in this work. 

We use the pre-trained network as a starting point for A3C. The cumulative reward over the A3C training period is presented in Fig. \ref{fig:plots}. As may be observed, pre-training the network helps to learn faster (2-3x) compared to learning from randomly initialized weights. We also experimented the FULL version of BiMGame with an additional reward as in Eqn. \ref{reward}, but with $d(s_t)$ being the geodesic distance to the center of the board instead of the radial distance. This reward function (even with different scaling factors) did not help to speed-up the learning process. We argue that this is a possible reason behind the value-based reward not playing a major role in faster learning.

It may be noted that we observed some unstable behavior while training STG2, resulting in high variations in time to converge compared to Full and STG1. For the sake of time, we report the best results obtained over a few runs of each algorithm. However, we noted at least 2x improvement with pre-training, in spite of such variations.

\section{Challenges and Future Work}
\label{sec:future}

BiMGame is a challenging task for both model-based, and model-free RL. In this work, we show that supervised pre-training on trajectories obtained with a simulator can speed up the learning. Yet the number of steps required is still in the order of millions. A combination of model-based and model-free RL seem to be necessary to further reduce the required number of steps, but for BiMGame further research is needed on how to achieve this.

We noticed that fine-tuning for game instances with more sparse rewards such as STG3 and STG4, seem to gradually forget its previous training, as it starts off with a small positive cumulative reward, which decreases gradually until it reaches zero. After many steps, none of the algorithms including randomly initialized A3C showed any sign of starting to learn these two tasks. Currently, it is an open problem. 
However, DAgger which entirely uses supervised training to mimic the MPC expert shows a very slow growth in cumulative reward, but requiring quite expensive query-to-expert.

Humans can learn to play BiMGame without much effort, particularly different game instances, i.e. different geometry, material, number of marbles. An interesting research direction may be to involve human priors within the learning framework. To that end, breaking the task into a sequence of subtasks, which are determined automatically or from trajectories, seems desirable.

\bibliographystyle{abbrvnat}
\bibliography{bibliography}



\end{document}